%% file: main.tex
\documentclass[10pt,twocolumn,letterpaper]{article}

\usepackage{cvpr}
\usepackage{times}
\usepackage{epsfig}
\usepackage{graphicx}
\usepackage{amsmath}
\usepackage{amssymb}
\usepackage{color}
\usepackage{colortbl}
\usepackage[table]{xcolor}
\usepackage{caption}

\newcommand\blfootnote[1]{%
  \begingroup
  \renewcommand\thefootnote{}\footnote{#1}%
  \addtocounter{footnote}{-1}%
  \endgroup
}

\newcommand{\norm}[1]{\left\lVert#1\right\rVert}

\newcommand{\afterfig}{\vspace{-0.22cm}}

\usepackage[pagebackref=true,breaklinks=true,letterpaper=true,colorlinks,bookmarks=false]{hyperref}

\cvprfinalcopy 

\def\cvprPaperID{6553} 

\ifcvprfinal\fi

\input{defs}
\begin{document}

\title{\vspace{-1.5cm} Semantic Pyramid for Image Generation \vspace{-0.2cm}}

\author{Assaf Shocher$^{*1,2}$
\qquad
Yossi Gandelsman$^{*1}$
\qquad
Inbar Mosseri$^1$
\qquad
Michal Yarom$^1$
\\
Michal Irani$^{1,2}$
\qquad
William T. Freeman$^1$
\qquad
Tali Dekel$^1$ \vspace{0.2cm}
\\
\small {$^1$ Google Research \qquad \qquad \qquad} $^2$Weizmann Institute of Science
}
\twocolumn[{%
\renewcommand\twocolumn[2][]{#1}%
\maketitle%
\vspace{-0.5cm}%
\input{teaser.tex}%
}]
\blfootnote{$^*$ indicates equal contributions.}
\blfootnote{The first author performed this work as an intern at Google.}

\input{abstract_new.tex}
\input{introduction.tex}

\begin{figure*}
    \centering
    \includegraphics[width=1\textwidth]{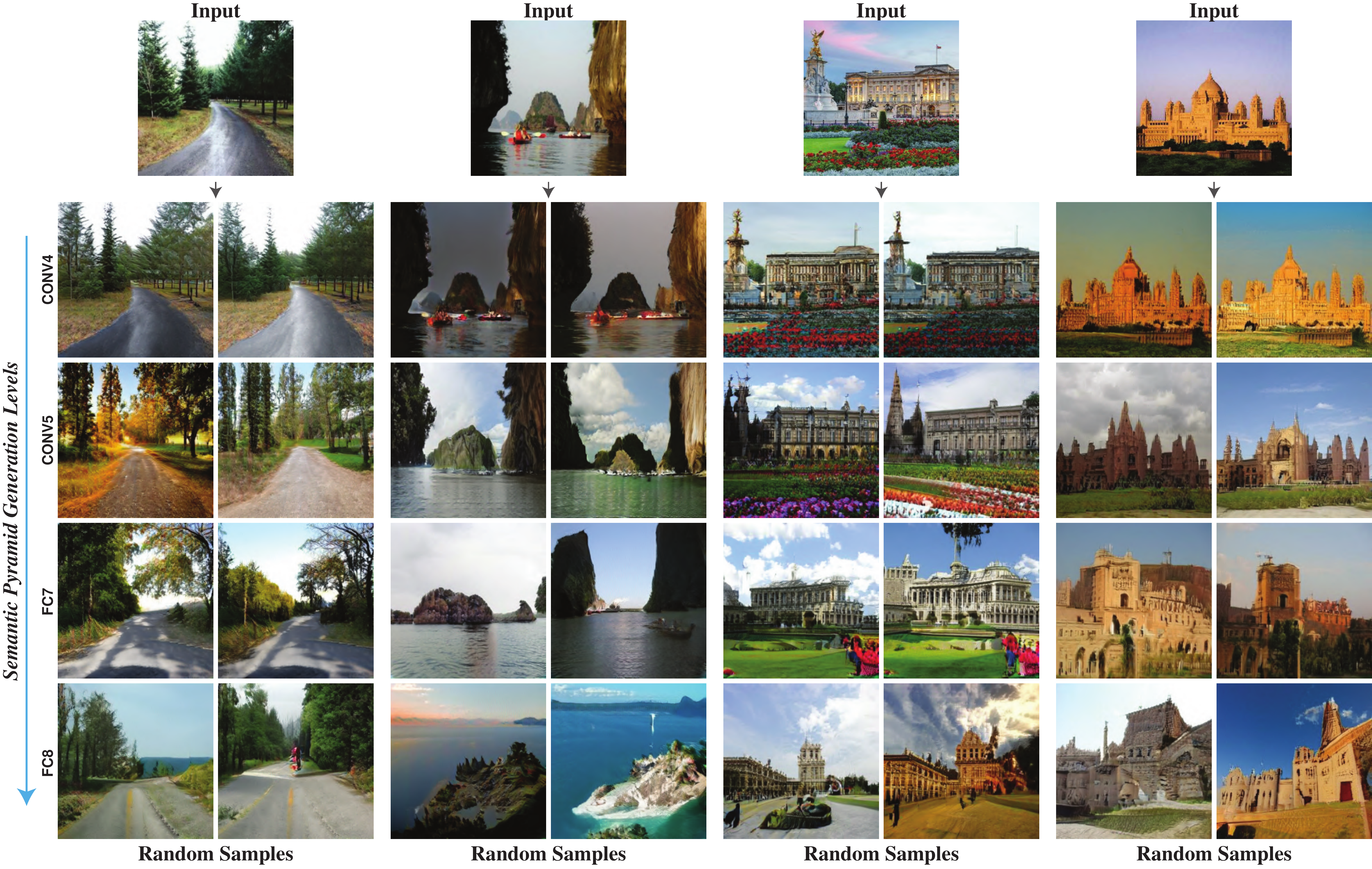}
    \caption{{\bf Random image samples from increasing semantic levels.} For each input image (top row), we show random image samples generated by our model when fed with deep features from increasing semantic levels (top to bottom). Our model generates realistic image samples, where the similarity to the original image is controlled by the level of the input features --- the deeper the features' level is, the more we can deviate from the original image. For all generation levels, our model preserves the semantic content of the input image and produces high quality diverse image samples.}
    \label{fig:inversion}\afterfig
\end{figure*}

\begin{figure*}[t!]
    \centering
    \includegraphics[width=\textwidth]{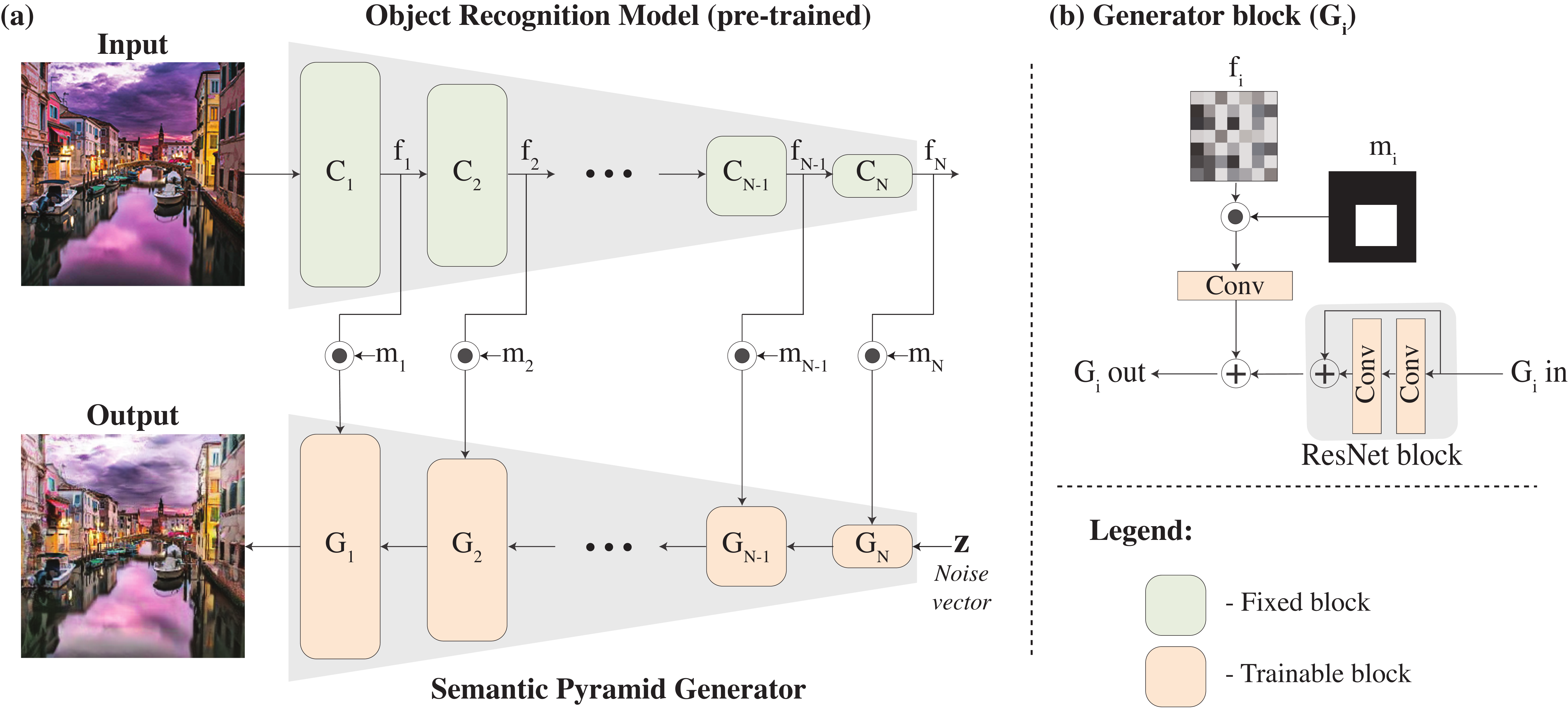}
    \caption{{\bf Semantic pyramid image pipeline.} (a) The generator works in full mirror-like conjunction with a pre-trained classification model. Each stage of the classification model has a corresponding block in the generator. (b) Specification of a single generator block. the feature map is first multiplied by its input mask. The masked feature map then undergoes a convolution layer and the result is summed with the result of the corresponding generator block.}
    \label{fig:pipeline}\afterfig
\end{figure*}

\begin{figure}
    \centering
    \includegraphics[width=1.0\columnwidth]{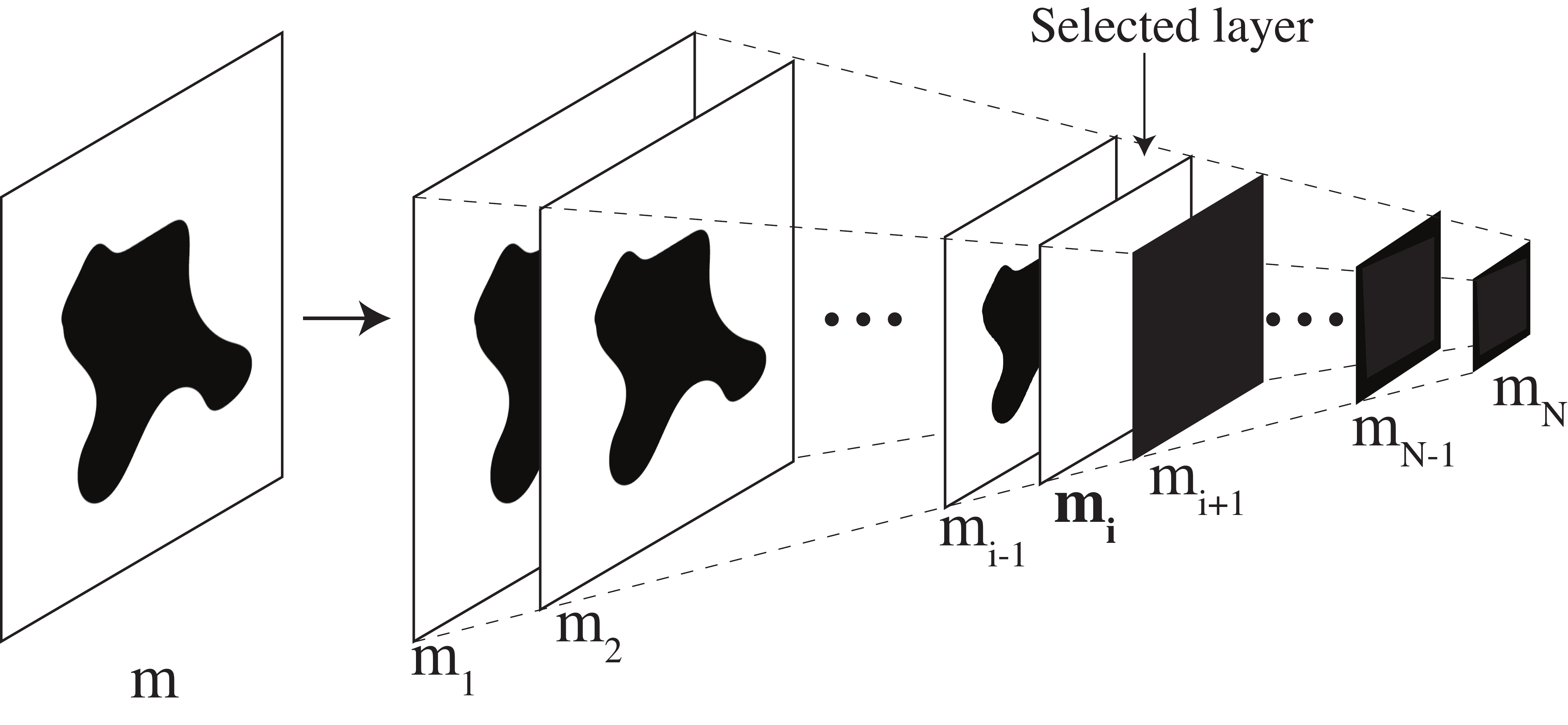}
    \caption{{\bf Applying spatially varying masks.} To generate only wanted areas of the image, feature maps are multiplied with masks. White indicates ``pass" and black indicates ``block". For training, a random blocked crop is sampled as well as a random ``selected layer". at inference time, a user can set any shape of the mask and determine the "selected layer" according to the extent of divergence wanted w.r.t to original input.}\afterfig
    \label{fig:mask_gen}\afterfig
\end{figure}

\begin{figure}
\hspace{-0.7cm}
    \centering
    \includegraphics[width=1.07\columnwidth]{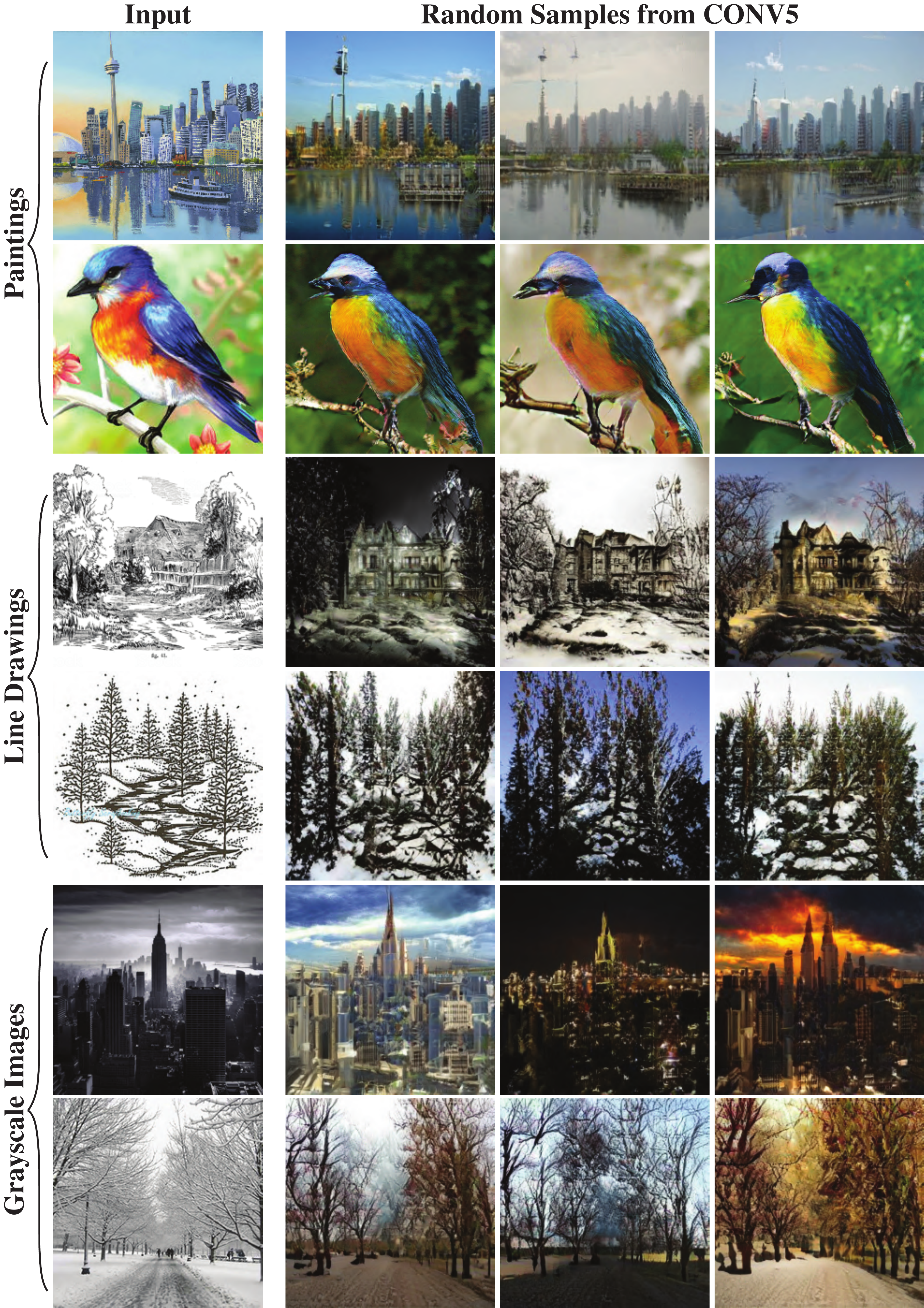}
    \caption{{\bf Image generation from paintings, line drawings and grayscale images}. Our model produces high quality, diverse image samples even when the input features are extracted from unnatural images such as painting or line-drawings, or images that are scarce in the training data such as grayscale images. In all these cases, our generated image samples convey realistic  image properties that do not exist in the original input images, including texture, lighting and colors.
  }
    \label{fig:out_of_dist}\afterfig
\end{figure}

\input{background.tex}

\begin{figure}
\hspace{-0.7cm}
    \centering
    \includegraphics[width=1.02\columnwidth]{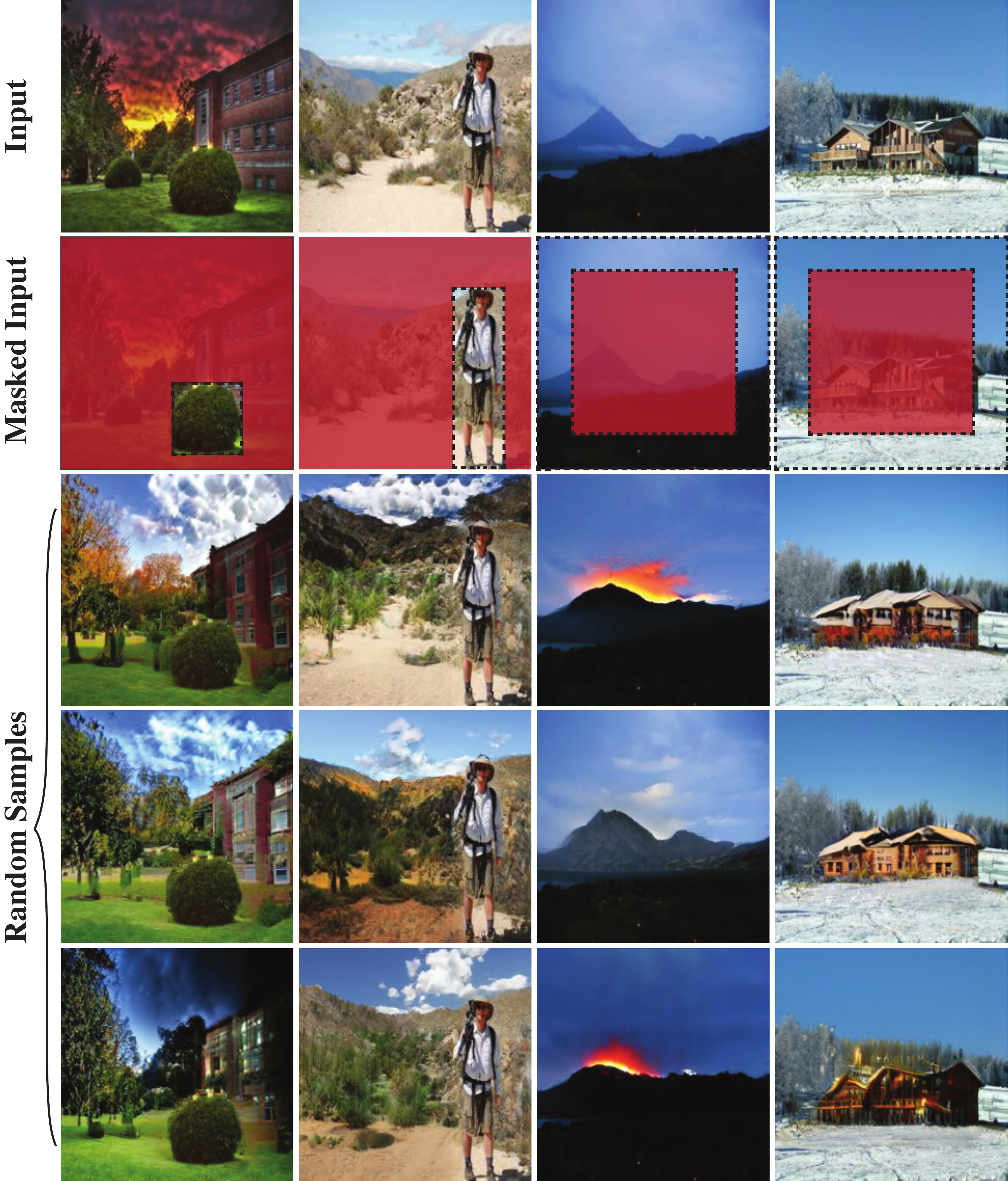}
    \caption{{\bf Image Re-painting.} We use our model to generate new random samples for desired \emph{image regions}, marked in red (second row) over the original images (first row); the image content in the unmasked regions is persevered. Our model generates diverse region samples that match the semantic content of the original image, and blends them naturally with the unmasked regions which remain intact.}
    \vspace{-0.5cm}
    \label{fig:repainting}
\end{figure}

\input{method.tex}

\begin{figure}
\hspace{-0.7cm}
    \centering
    \includegraphics[width=1.07\columnwidth]{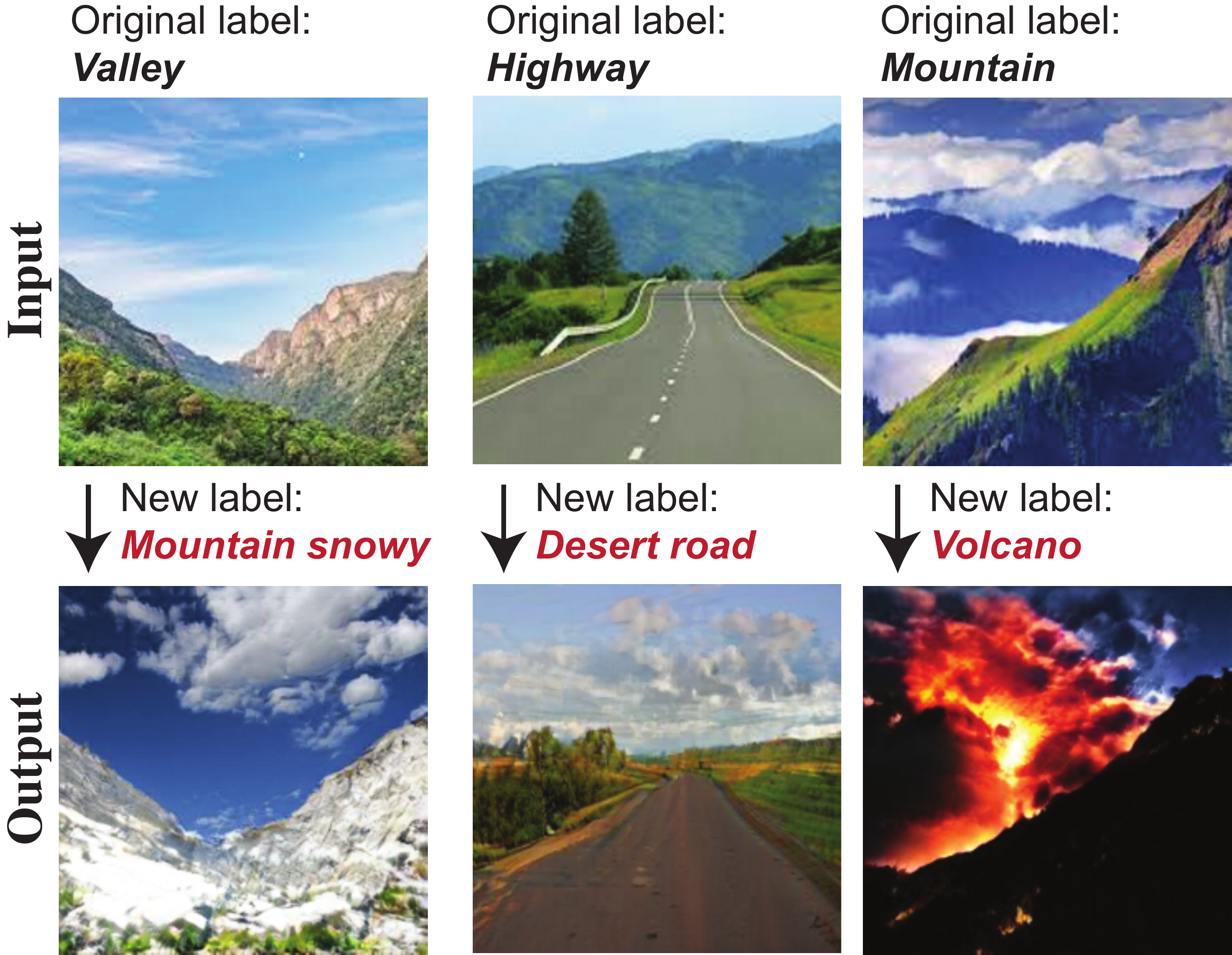}
    \caption{{\bf Image re-labeling.} Given an input image (top row), with its class label (estimated by the classification network), we generate a new image by feeding to our model its original mid-level features, yet modifying its class label, e.g., \emph{Valley}$\rightarrow$\emph{Mountain snowy} (see Sec.~\ref{sec:results}). By doing so, we can change the semantic properties of the image, while preserving its dominant structures. }
    \label{fig:relabel}
\vspace{-2em}
\end{figure}

\input{results}

\input{conclusion.tex}

\clearpage
{\small
\bibliographystyle{ieee_fullname}
\bibliography{refs}
}

\end{document}

%% file: teaser.tex
\centering
	\centering
\includegraphics[width=0.90\textwidth]{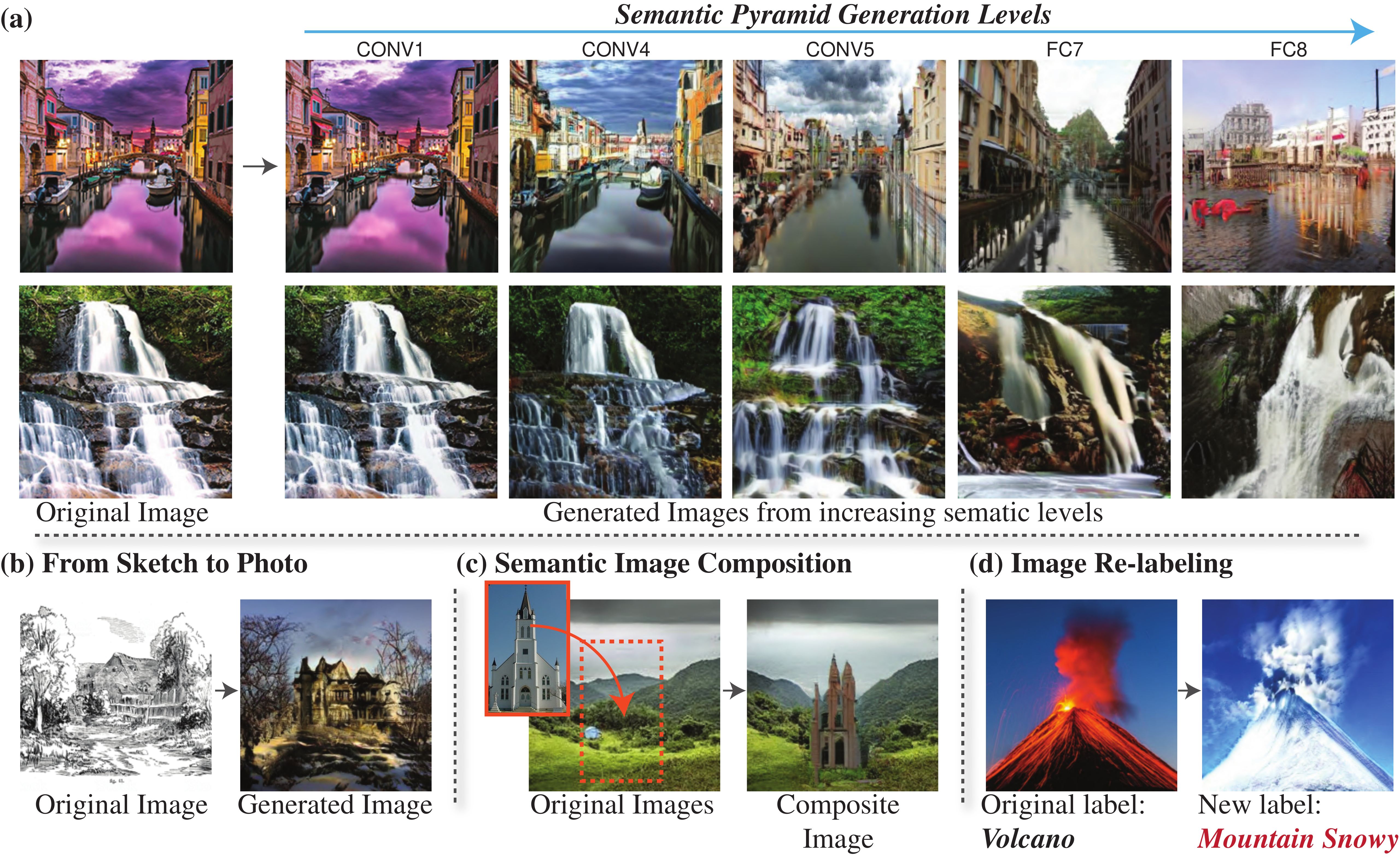}\vspace{-0.1cm}
\captionof{figure}{{\bf Semantic Pyramid for Image Generation.} We introduce a new image generative model that is designed and trained to leverage the hierarchical space of deep-features learned by a pre-trained classification network. Our model provides a unified versatile framework for various image generation and manipulation tasks, including: (a) generating images with a controllable extent of semantic similarity to a reference image, obtained by reconstructing images from different layers of a classification model; (b) generating realistic image samples from unnatural reference image such as line drawings; (c) semantically compositing different images, and (d) controlling the semantic content of an image by enforcing a new, modified class label.}
\afterfig
\label{fig:teaser}
\vspace{0.35cm}


%% file: abstract_new.tex
\begin{abstract}
We present a novel GAN-based model that utilizes the space of deep features  learned by a pre-trained classification model.  Inspired by classical image pyramid representations, we construct our model as a Semantic Generation Pyramid -- a hierarchical framework which leverages the continuum of semantic information encapsulated in such deep features; this ranges from low level information contained in fine features to high level, semantic information contained in deeper features.  More specifically, given a set of features extracted from a reference image, our model generates diverse  image samples, each with matching features at each semantic level of the classification model.  We demonstrate that our model results in a versatile and flexible framework that can be used in various classic and novel image generation tasks. These include: generating images with a controllable extent of semantic similarity to a reference image, and different manipulation tasks such as semantically-controlled inpainting and compositing; all achieved with the same  model, with no further training.
\footnote{Project website: \url{https://semantic-pyramid.github.io/}}
\end{abstract}

%% file: introduction.tex
\section{Introduction}

Convolutional Neural Networks (CNNs) trained for visual classification were shown to learn powerful and meaningful feature spaces, encoding rich information ranging from low level features to high-level semantic content~\cite{alexnet}. Such features have been widely utilized by numerous methods for clustering, perceptual loss \cite{zhang2018perceptual} and different image manipulation tasks  \cite{Upchurch_2017, style_transfer, Johnson_2016, arar2019image}.  

 The process of working in feature space typically involves the following stages: an image is fed to a pre-trained classification network; its feature responses from different layers are extracted, and optionally  manipulated according to the application at hand. The manipulated features are then inverted back to  an image  by solving a reconstruction optimization problem. However, the problem of inverting deep features into a realistic image is challenging -- there is no one-to-one mapping between the deep features and an image, especially when the features are taken from deep layers. This has been addressed so far mostly by imposing regularization priors on the reconstructed image, which often leads to blurry unrealistic reconstructions and limits the type of features that can be used.

In this paper, to overcome the aforementioned limitations, we take the task of feature inversion to the realm of Generative Adversarial Networks (GANs). GANs have made a dramatic leap in modeling the distribution of natural images and are now capable of generating impressive realistic image samples. However,  most existing GAN-based models that use deep features condition the generation only on objects' class information~\cite{acgan, sngan, SAGAN, BIGGAN}. In contrast, we propose a novel generative model that \emph{utilizes the continuum of semantic information encapsulated in deep features}; this ranges from low level information contained in fine features to high level, semantic information contained in deeper features. By doing so, we bridge the gap between optimization based methods for feature inversion and generative adversarial learning.

Inspired by classical image pyramid representations, we construct our model as a \emph{Semantic Generation Pyramid} -- a hierarchical GAN-based framework which can leverage information from different semantic levels of the features. More specifically, given a set of features extracted from a reference image, our model can generate diverse  image samples, each with matching features at each semantic level of the classification model. This allows us to generate
images with a gradual, controllable semantic similarity to a
reference image (see Fig.~\ref{fig:teaser} and Fig.~\ref{fig:inversion}). 

The hierarchical nature of our model provides a versatile and flexible framework that can be used for a variety of \emph{semantically-aware image generation and manipulation tasks}.  Similar to classic image pyramid representations, this is done by manipulating features at different semantic levels, and controlling the pyramid levels in which features are fed to our model. We demonstrate this approach in a number of applications, including semantically-controlled inpainting, semantic compositing of objects from different images, and generating realistic images from grayscale, line-drawings or paintings. 
All these tasks are performed with the same unified framework, without any further optimization or fine tuning.

%% file: background.tex
\section{Related Work}

\paragraph{Inverting deep features.} Inverting deep features back to images has been mostly studies in the context of interpretability and understanding of visual recognition networks.  Simonyan \emph{et al.}~\cite{journals/corr/SimonyanVZ13} formulated the feature inversion problem as an optimization problem  where the objective is to minimize the $L_2$ distance between the mapping of the image to features (by the pre-trained network) and a given feature map. They  apply back-propagation to minimize this objective -- a slow process, which is highly sensitive to initialization. An important observation made by this process was how reconstructable an image is from various depths of CNN layers; first several layers are almost fully invertible, but the ability to reconstruct the input image quickly declines along the depth of the network. 

Other optimization-based methods attempt to the reconstruction from deeper features by imposing different regularization priors on the reconstructed image \cite{Mahendran_2015, nguyen2016, journals/corr/YosinskiCNFL15}. However, these methods are able to reconstruct only a single, average image. Because there is no one-to-one mapping between deep features to an image, the reconstructed image is often blurry and unrealistic.


Dosovitskiy\&Brox \cite{DBLP:journals/corr/DosovitskiyB15} propose to train a CNN to invert various image descriptors, among which are deep features. This approach also impose  regularization, yet implicitly; the fact that an image was generated by a CNN forms a strong natural image prior \cite{DIP, DoubleDIP}. However,  such regularization also tends to produce blurry unrealistic images when inverting the deeper features. 

To overcome this limitation, a generative model for feature invertion was proposed by the same authors~\cite{Dosovitskiy:2016:GIP:3157096.3157170}. 
Although they used a GAN, their model is still deterministic, i.e., generates only a single possible image from an input set of features. In this case the discriminator was used as a learned regularizer to try to impose realistic reconstructions from the generator. This resulted in images that have local structure that looks natural but distorted unrealistic global structure. The final image cannot be considered realistic (See examples in supplementary material). The authors mention a desire to add diversity and randomization to their process and report an attempt to inject noise for this purpose. This attempt failed due to the fact that "Nothing in the loss function forces the generator to output multiple different reconstructions". We address this problem using a \emph{diversity-loss}.
\vspace{-0.4cm}
\paragraph{ Deep features for image manipulation.} Inverting deep features crossed the line from the field of interpretability and understanding to image manipulation. A general approach is to apply some manipulation to semantic features, and then invert back to pixels to project the manipulation to a resulting output image. Such manipulation include Texture-synthesis \cite{texturesynth}, Style-transfer \cite{style_transfer}, feature interpolation, demonstrated to apply aging to faces \cite{Upchurch_2017} and recently also image retargeting by applying Seam Carving \cite{seamcarving} to semantic features \cite{arar2019image}. In all of these works the output image is reconstructed by some variant of the iterative optimization process of \cite{journals/corr/SimonyanVZ13} that takes time and is sensitive to initialization. Some solutions to speed up were proposed, like training a CNN to imitate the mapping of the optimization process \cite{Johnson_2016}.

\vspace{-0.3cm}
\paragraph{Generative Adversarial Networks} In our work we make use of recent advances of Generative Adversarial Networks (GANs) \cite{gan}. There has been huge progress in the quality of image generation using GANs \cite{dcgan, acgan, sngan, SAGAN,BIGGAN}. Our GAN is based on Self-Attention GAN \cite{SAGAN} with slight modifications. Differently from classical GANs, we perform image to image mapping using a conditional GAN similarly to \cite{pix2pix2017,pix2pixHD}. There has been some recent impressive work on interpreting and controlling the outcomes of GANs by either interpreting neurons \cite{bau2019gandissect} or by steering the latent space \cite{gansteerability}. Our approach differs due to the use of semantic features that originate in supervised classification networks. \cite{huang2017sgan} makes use of classification feature maps to improve quality of classic generation tasks. They stack a set of GANs first trained separately to generate features of different levels and then combine them. We have different goal and setting of how to make use of the semantic features. Further analysis \cite{bau2019seeing} has shown limitations on what GANs cannot generate. We introduce the application of Re-painting which allows regenerating selected parts of the image and by that enables keeping wanted parts of an image as is. To some practical uses, this overcomes the limitations presented in \cite{bau2019seeing} (such as inability to generate humans).
\vspace{-0.2cm}
\paragraph{Classical hierarchical image representations.} We draw inspiration to our method from the classical image processing approach of image pyramids, especially Laplacian pyramids \cite{Burt83thelaplacian, adelson1984pmi}. This method decomposes images to distinct frequency-bands thus allows frequency aware manipulation for stitching and fusion of images. reconstruction is fast and trivial. Our method is a semantic analogous to this approach. We aim to perform semantic manipulation and get immediate projection back to the image pixels.

%% file: method.tex
\section{Method}
Our goal is to design a generative image model that can fully leverage the feature space learned by a pre-trained classification network. More specifically, we opt to achieve the following objectives:

\begin{enumerate}
\item \emph{Leveraging features from different semantic levels.} Given an input image, the deep features extracted from different layers have hierarchical structure -- features extracted from finer layers of the model contain low level image information, while deeper features can encode higher level, semantic information \cite{yosinski-understanding}.  We would like to benefit from the continuum space of these features. 
    \item \emph{Flexibility and user controllability.} We want to support various manipulation tasks at test time via editing in the deep feature space. For example, combining features from different images, or from different levels.  The model then have to provide such user controllability and adapt to various manipulations for the features. 
    
    \item \emph{Diversity.} We would like our model to learn the space  of possible images that match a given set of input features, rather than producing a single image sample. 
\end{enumerate}

We next describe how we achieve these objectives via a unified  GAN-based architecture and a dedicated training scheme.

\subsection{Architecture}
Our generator works in full conjunction with a a pre-trained classification model, which we assume is given and fixed. In practice, we use VGG-16 model \cite{vgg} trained on Places365 dataset \cite{places} dataset. More specifically, given an input image $x$, we feed it into the classification model and extract a set of features maps  $\mathcal{F}=\{f_l\}$  by taking the activation maps of different layers of the model. That is, $f_l = C^*_l(x)$, where $C^*_l$ dnotes the $l$-th layer of the classification model. These features are then fused into our generator as follows.

Our generator's architecture is loosely based on the class-conditioned GAN~\cite{SAGAN}. However, we modify it to have a  mirror structure w.r.t. the classification model, as illustrated in Fig.~\ref{fig:pipeline}. More specifically, each residual-block in our generator corresponds to a single stage in the classification model (a stage consists of 2-3 conv layers + pooling). This structure forms a \emph{semantic generation pyramid}, which at the coarsest level takes a random noise vector, $z$, as input. At each of the upper levels, our model can optionally receive features, $f_l$, extracted from the corresponding level of the classification model. The flow of the features from the classification model to our generator is controlled at each level by an input mask $m_l$. The mask can either pass the entire feature map (all ones), mask-out the entire feature map (all zeros) or pass regions from it. 

To conclude, the input to the network is: (1) a set of deep features, $\mathcal{F}=\{f_l\}$  computed by feeding an input image, $x$, into the classification model and extracting the activation maps from different layers; (2) a noise vector, $z$, which allows diversity and learning a distribution rather than a one-to-one mapping; (3) a set of masks $\mathcal{M}=\{m_l\}$, each for input feature $f_l$; these masks allows us to control, manipulate and leverage features from different semantic levels. Thus, the generator can be formulated by
$ G(z,\mathcal{F},\mathcal{M}) $.

Fig.~\ref{fig:pipeline}(b) depicts how the feature maps are fused into our generator. The goal is to combine information both from the current classification model layer and from previous generator blocks that originate in the noise vector. At each level, the feature map $f_l$ is first multiplied by its input mask $m_l$. The masked feature map  then undergoes a convolution layer and the result is summed with the result of the corresponding generator block. In cases where the entire feature map is masked, nothing is added to the result of the previous generator block. The mask itself is concatenated as another channel to allow the proceeding layers awareness and distinction between masked areas and  empty areas. 

As in \cite{SAGAN}, the generator consists of residual-blocks \cite{resnet}. We used self-attention layers in both the generator and discriminator. The discriminator is the same as \cite{SAGAN}.



\begin{figure}
\hspace{-0.7cm}
    \centering
    \includegraphics[width=1.07\columnwidth]{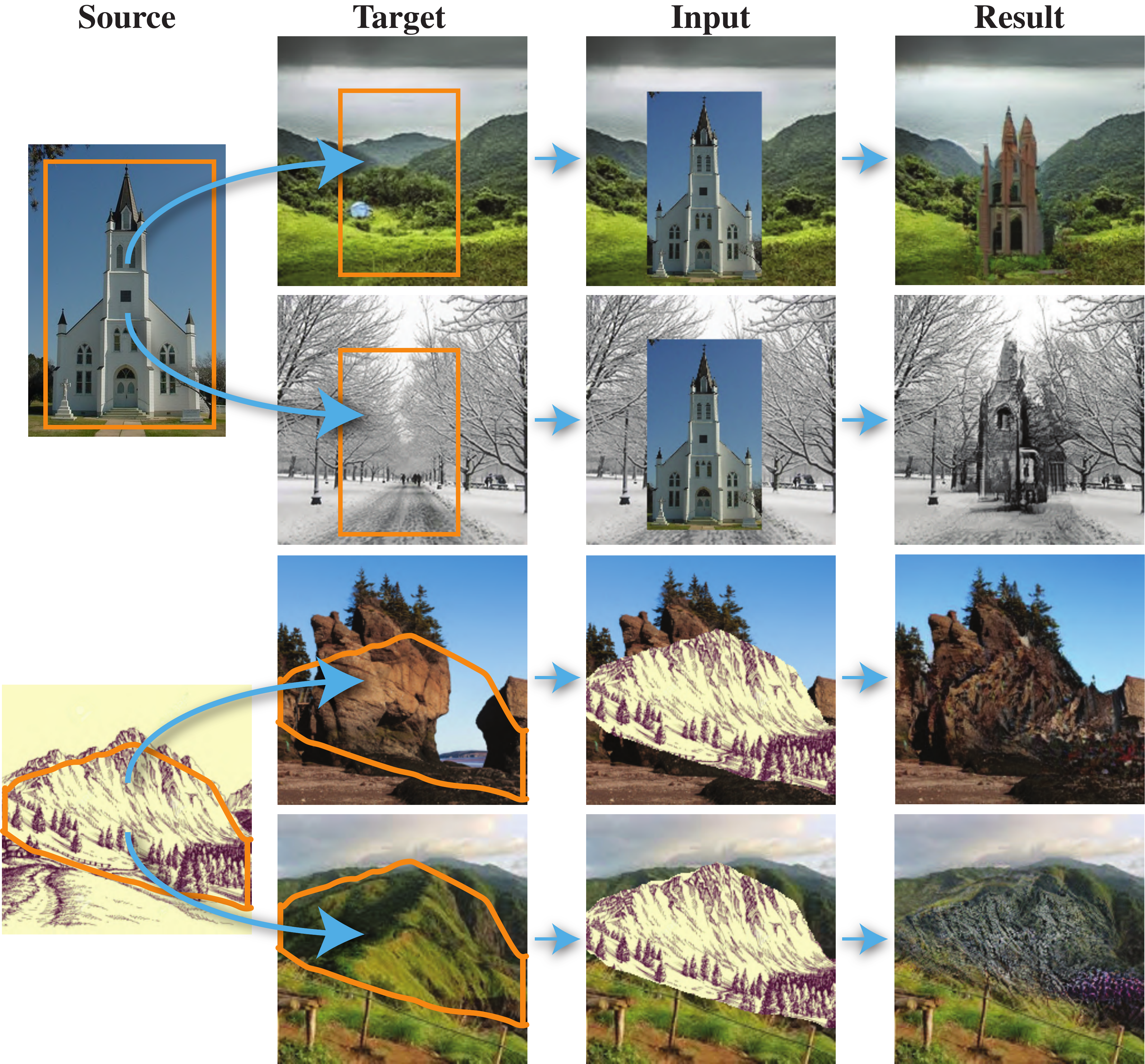}
    \caption{{\bf Semantic Image Composition.} Implanting an object or some image crop inside another image, such that the implanted object can change in order to fit its surroundings but still hold on to its semantic interpretation.}
    \label{fig:harmonization}\afterfig
\end{figure}

\subsection{Training procedure}
Our goal is to have a single unified model that can generate images from any level of our semantic pyramid. In other words, we want to be able to generate diverse, high quality image samples from any \emph{subset} of input features $\{f_i\} \subset \mathcal{F}$. We achieve this through the following training scheme.

At each training iteration, a batch of input images is sampled from the dataset and is being fed to classification model to compute their features.  In our default training step, we randomly select a pyramid level, and feed to the generator only the features at that level, while masking out the features in all other levels.

However, we also want the ability to generate from a mixture of semantic levels, keeping some areas of the image while modifying other areas. We therefore train with spatially varying masks too. At some of the iterations (defined by a hyper-parameter probability), we incorporate spatially varying masks. Fig.~\ref{fig:mask_gen} depicts the masking procedure in such cases. First, random crop of the image is sampled. Then, for one randomly selected layer the mask is fully turned on. For all the other layers, closer to the input image, the mask is turned on except the sampled crop. This sort of training is oriented towards tasks of editing different spatial areas of the image differently.

\subsection{Losses}
We train all the levels in our pyramid architecture simultaneously, where our training loss consists of three terms given by:
\afterfig
\vspace{-0.15cm}
\begin{equation}
\vspace{-0.15cm}
\min_{G}\,\,\max_{D}\, \mathcal{L}_{\texttt{adv}}(G, D) +\alpha\mathcal{L}_{\texttt{rec}}(G)+\beta\mathcal{L}_{\texttt{div}}(G)
\label{eq:loss}
\vspace{-0.15cm}
\end{equation}
\vspace{-0.15cm}

The first term $\mathcal{L}_{\texttt{adv}}$ is an adversarial loss. That is, our generator is trained against a class-conditioned discriminator $D$, similar to~\cite{SAGAN}. We adopt the LSGAN~\cite{lsgan} variant of this optimization problem. Formally, 
\vspace{-0.1cm}\begin{equation}
\begin{split}
\hspace*{-0.3cm}
& \mathcal{L}_{\texttt{adv}}(G,D) = \\
& \mathbb{E}_{x \sim p_{\text{data}}(x)}[(D(x)-1)^2]+\\
& \mathbb{E}_{z \sim p_z(z), \mathcal{F} \sim p_{\text{data}}(\mathcal{F}), \mathcal{M} \sim p_m(\mathcal{M})}[D(G(z,\mathcal{F}, \mathcal{M}))^2]
\end{split}
\end{equation}
Where $ p_z $ is normal distribution for noise instances, and $ p_m$ is the distribution for sampling masks as described above.

The second term $\mathcal{L}_{\texttt{rec}}$ is a \emph{semantic reconstruction loss}, \emph{which encourages the output image to preserve the feature information used to generate it.} This is similar to perceptual loss \cite{Johnson_2016, zhang2018perceptual}. More specifically, when feeding a generated image back to the classification model, we want to ensure that the resulting features will be as close as possible to the features extracted from the original image. To allow the model to generate diverse image samples from high level features, we apply this loss only to the features at levels that are used for generation (not masked out). Formally,


    \vspace{-0.15cm}
\begin{equation}
    \vspace{-0.15cm}
\mathcal{L}_{\texttt{rec}} = \sum_{l \in \text{layers}}{\norm{\left(C^*_l(x)-C^*_l(G(z,\mathcal{F},\mathcal{M})\right) \cdot m_l}_1}
    \vspace{-0.1cm}
\end{equation}
    \vspace{-0.1cm}

Where $C_l$ denotes the $l$ layer of the classification model.  Both the original feature maps and the reconstructed ones are normalized together so that comparison is agnostic to global color scaling. Furthermore, to allow more geometric diversity and not imposing pixel to pixel complete matching, we first apply max-pooling to both original and reconstructed feature maps, with $2\times2$ windows grid. So we are effectively only comparing the strongest activation in each window, allowing slight shifts in locations (which translate to bigger shifts in image pixels the deeper the feature map is).

Finally, $\mathcal{L}_{\texttt{div}}$ is a diversity loss, based on \cite{MSGAN}. Specifically, each batch is divided into pairs of instances that have the same input image and masking (but different noise vectors). A regularization is applied such that the $L_1$ distance between two such generated results should be higher as the two noise vectors are distant one from the other. 
\begin{equation}
\mathcal{L}_{\texttt{div}} = \frac{\norm{z_1 - z_2}_1}{\norm{G(z_1,\mathcal{F},\mathcal{M}) - G(z_2,\mathcal{F},\mathcal{M})}_1 + \epsilon}
\vspace{-0.15cm}
\end{equation}
\vspace{-0.15cm}

%% file: results.tex
\subsection{Implementation details}
We use VGG-16 \cite{vgg}. The inputs to the generator are the feautres at the end of each stage (after the pooling layer). We also use the fully-connected layers, FC7 and FC8.
To enable matching between the SA-GAN generator to the VGG classifier, we did not use FC6. We trained our model on Places365 dataset \cite{places}. We used the loss as indicated in Eq.~\ref{eq:loss} with $\alpha=0.1$ and $\beta=0.1$. The probability for training with missing crops was set to $0.3$. We used Tensorflow platform with TFGAN tool.
We trained the model for approximately two days on Dragonfish TPU with $4\times4$ topology. We employed some methods from \cite{BIGGAN} such as class-embeddings and truncation trick. Optimization was by Adam optimizer \cite{adam} with learning rate of $10^{-4}$ for both the generator and discriminator. Batch size of 1024 and 128 latent space dimensions as in \cite{SAGAN}.

\section{Experiments and Applications}
\label{sec:results}


We tested our model on various natural images taken from Places365 \cite{places} and downloaded from the Web. 
Fig.~\ref{fig:teaser} and Fig.~\ref{fig:inversion} show several qualitative  results of our generated images from increasing semantic levels. That is, for each pyramid level, we feed our generator with features extracted  only from that level. In Fig.~\ref{fig:inversion}, we show for each example and for each semantic level two different random  image samples, generated both from the same features, but with different noise instances. 


As can be seen, the pyramid level from which the image is generated determines to which extent it diverges from the reference image -- the fidelity to the reference image decreases with the pyramid level, whereas the diversity between different image samples increases. 
Nevertheless, for all generation levels, the semantic content of the original image is preserved.

Observing our generated images more closely reveals details about the distribution of images matching to a feature map at each level. For example, it is apparent that CONV4 layer is agnostic to mild lighting and color modifications but preserves geometric structures and textures. The fully connected layers are mostly agnostic to geometric global structure but preserve textures and local structures (e.g., bottom row of Fig.~\ref{fig:inversion}); the global shape of the road, the position of the island and the architecture of the castles has completely changed. However, local attributes such as the existence of small windows, towers remained (even though in different locations). This aligns with observations made by \cite{noglobal}.



\subsection{Quantitative Evaluation}
We evaluated the quality of our generated images using two measures; Fr\'echet Inception Distance (FID) \cite{Heusel} and ``Real/Fake'' user study. 

For FID, we used 6000 test images randomly sampled from Places365~\cite{places}. We extracted deep features by feeding these images to the classification model, and then generated random image samples from each semantic level separately, i.e., by inputting to our model only the features extracted from that level.  

Table.~\ref{fid} reports FID scores  measured for our generation results from each semantic level. As expected, the finer the features' level is the lower the FID score. For example, when generating images from CONV1 features the distribution of generated images almost perfectly aligns with the real images. As the features' level increase, our generated images  deviate more from the original images, which is reflected by a consistent increase in the FID scores.

For the user study, we used Amazon Mechanical Turk (AMT), following the protocol of \cite{zhang2016colorful, pix2pix2017}. The following two tests were given: \vspace{-0.15cm}
\begin{enumerate}
    \vspace{-0.07cm}
    \item  Paired test: A generated image is presented against its corresponding reference image (i.e., the features used for generation are extracted from the reference). The workers were asked to select the fake one.
    \vspace{-0.25cm}
    \item Unpaired test: A generated image is presented against some real unrelated image. The workers were asked asked to determine whether it is fake. 
        \vspace{-0.15cm}
\end{enumerate}
    In each trail,  images were presented  for 1 sec. Each of these tests was performed by 100 raters, using 75 images randomly sampled from Places365 \cite{places}; 
    to prevent immediate discrimination between real and fake images, we did not include images with people in this test (generally, humans are not synthesized well by GANs \cite{bau2019seeing}, as well as by our model when fed with features from high semantic levels).

Table.~\ref{amt} reports the confusion rate (percentage of fooled turkers) separately measured for  images generated from each semantic level. Perfect confusion is 50\%. This means, for example, that the images generated from CONV1 are almost indistinguishable from real ones. Similarly, About 17\%-18\% of the images generated from FC8 looked more genuine than the real ones shown.

\begin{table}
\resizebox{\columnwidth}{!}{%
\begin{tabular}{|l|l|l|l|l|l|l|l|}
\hline
    Conv1  & Conv2  & Conv3  & Conv4  & Conv5  & FC7  & FC8 \\ \hline

2.89 & 8.67 & 11.08 & 17.64 & 19.59 & 22.67 & 29.34   \\
\hline
\end{tabular}
}
\caption{{\bf FID per semantic level}. In each column, we report the estimated FID score when our  images are generated from different semantic levels. As expected, the finer the level is the lower the FID score is.}
\vspace{-0.3cm}
\label{fid}
\end{table}

\begin{table}
\resizebox{\columnwidth}{!}{%
\begin{tabular}{|l|l|l|l|l|l|l|l|}
\hline
    & Conv1  & Conv2  & Conv3  & Conv4  & Conv5  & FC7  & FC8 \\ 
    \hline
Paired & 49.6\% & 42.7\% & 22.2\% & 20.9\% & 16\% & 19.1\% & 18\% \\
\hline
Unpaired & 51.1\% & 39.1\% & 27.6\% & 15.1\% & 13.3\% & 21.6\% & 17.2\% \\
\hline
\end{tabular}
}
\caption{{\bf AMT Real/Fake user study:} We report confusion rates ($\%$ of fooled turkers) for two types of tests: (i) Paired: generated image against its reference image, and (ii) Unpaired:generated image, against some real unrelated image. We report the results for images generated from different pyramid levels.
}
\vspace{-0.3cm}
\label{amt}
\end{table}


\subsection{Semantic Pyramid Image Manipulation}

The pyramid structure of our model provides a flexible framework that can be used for various \emph{semantically-aware image generation and manipulation tasks}.  Similar to classic image pyramid representations~\cite{Burt83thelaplacian, adelson1984pmi}, this can be done by manipulating features at different semantic levels, and controlling the pyramid levels in which features are fed to our model. We next describe how this approach can be applied for a  number of applications.  Note that we use the same model, which was trained once and used at inference mode for all applications.


\vspace{-0.3cm}
\paragraph{Re-painting.}
We introduce a new application, we name  \emph{re-painting}, where  an image region can be re-generated, with a controllable semantic similarity to the original image content.  In contrast to  traditional  in-painting, in which there is no information in the generated region (e.g., \cite{Barnes:2009:PAR,yang2016highresolution,Teterwak_2019_ICCV}), we utilize the information available in deep features for that region.  In other words, re-painting allows us to re-sample the image content in a specific region based on its original content.

Fig.~\ref{fig:repainting} shows a number of re-painting results. As can be seen, our model successfully replace the desired regions with diverse generated region samples, while preserving the content around it.  This enables practical image manipulations such as generating the same hiker at various environments (Fig.~\ref{fig:repainting} second column from left), or replacing a house with various other houses while keeping the same environment (rightmost column).
These results demonstrate the ability of our network to fuse information from different levels at different image regions.  Since our training procedure of the network combines generating from different semantic levels (different layers of the classifier) within the same image, our generator can produce plausible images when some spatial piece of the image is generated from a more semantic feature-map. 

Fig.~\ref{fig:mask_gen} depicts how re-painting is done at both training and inference. The feature map matching the semantic level we want to repaint from is fed to the generator. Then we mask out the wanted re-painted region from all the feature-maps that are closer to the input image (the less semantic).

\vspace{-0.3cm}
\paragraph{Semantic image composition.}
The technique introduced for re-painting can be expanded and used for the challenge of Semantic Image Composition. Namely, implanting an object or some image crop inside another image, such that the implanted object can change in order to fit its surroundings but still hold on to its semantic interpretation. Fig.~\ref{fig:harmonization} shows such examples. Note how the church changes its structure and color according to its surroundings. These are semantic changes as oppose to just matching textures and lighting; In the top example the church does not just match by texture, it is transformed to a temple that is more likely to be found in such surrounding. Composition is done similarly to Re-painting; the only difference is that for the chosen most semantic layer we use a naive pasting of the object on the image. Then we mask out the matching region from all the feature-maps that are closer to the input image.

\vspace{-0.3cm}
\paragraph{Generation from unnatural reference image.}
Fig.~\ref{fig:out_of_dist} demonstrates the effect of using a reference image which does not belong to the distribution, i.e. not a a natural RGB image. Since the generator was trained to output images that belong to the dataset distribution, we get an image to image translation. We demonstrate converting paintings to realistic photos, line drawings to images and coloring grayscale images. For each case we generate a diverse set of possible results. Differently from \cite{pix2pix2017,pix2pixHD}, there is no exact matching of pixels between the reference image and the output. Inverting from CONV5 features has some degree of spatial freedom and allows modifying the structure. As an example, the rightmost colorization of the city replaced the tower with two towers.

\vspace{-0.3cm}
\paragraph{Re-labeling.}
We demonstrate a rather simple application using our semantic pyramid; we use CONV5 features from an input image but manually change the class label fed to the generator. Fig.~\ref{fig:relabel} shows the effect of such manipulation. Our GAN based on \cite{SAGAN} is class conditioned. The conditioning in the generator is enforced through the conditional batch norm layers as introduced in \cite{SAGAN}. This means that the difference between regular inversion and relabeling in the generator is carried out by just normalizing activations to different mean and variance.

%% file: conclusion.tex
\section{Discussion and Conclusion}
This work proposes a method to bridge the gap between semantic discriminative models and generative modelling. We demonstrated how our semantic generation pyramid can be used as  a unified and versatile framework for various image generation and manipulation tasks. Our framework also allows exploring the sub-space of images matching specific semantic feature-maps. We believe that projecting semantic modification back to the pixels realistically, is a key for future work that involves image manipulation or editing through the semantic domain. We hope this  work can guide and trigger further progress in utilizing semantic information in generative models.





%% file: main.bbl
\begin{thebibliography}{10}\itemsep=-1pt

\bibitem{adelson1984pmi}
E.~H. Adelson, C.~H. Anderson, J.~R. Bergen, P.~J. Burt, and J.~M. Ogden.
\newblock {1984, Pyramid methods in image processing}.
\newblock {\em RCA Engineer}, 29(6):33--41, 1984.

\bibitem{arar2019image}
Moab Arar, Dov Danon, Daniel Cohen-Or, and Ariel Shamir.
\newblock Image resizing by reconstruction from deep features, 2019.

\bibitem{seamcarving}
Shai Avidan and Ariel Shamir.
\newblock Seam carving for content-aware image resizing.
\newblock {\em ACM Trans. Graph.}, 26(3):10, 2007.

\bibitem{Barnes:2009:PAR}
Connelly Barnes, Eli Shechtman, Adam Finkelstein, and Dan~B Goldman.
\newblock {PatchMatch}: A randomized correspondence algorithm for structural
  image editing.
\newblock {\em ACM Transactions on Graphics (Proc. SIGGRAPH)}, 28(3), Aug.
  2009.

\bibitem{bau2019gandissect}
David Bau, Jun-Yan Zhu, Hendrik Strobelt, Bolei Zhou, Joshua~B. Tenenbaum,
  William~T. Freeman, and Antonio Torralba.
\newblock Gan dissection: Visualizing and understanding generative adversarial
  networks.
\newblock In {\em Proceedings of the International Conference on Learning
  Representations (ICLR)}, 2019.

\bibitem{bau2019seeing}
David Bau, Jun-Yan Zhu, Jonas Wulff, William Peebles, Hendrik Strobelt, Bolei
  Zhou, and Antonio Torralba.
\newblock Seeing what a gan cannot generate.
\newblock In {\em Proceedings of the International Conference Computer Vision
  (ICCV)}, 2019.

\bibitem{BIGGAN}
Andrew Brock, Jeff Donahue, and Karen Simonyan.
\newblock Large scale {GAN} training for high fidelity natural image synthesis.
\newblock {\em CoRR}, abs/1809.11096, 2018.

\bibitem{Burt83thelaplacian}
Peter~J. Burt and Edward~H. Adelson.
\newblock The laplacian pyramid as a compact image code.
\newblock {\em IEEE TRANSACTIONS ON COMMUNICATIONS}, 31:532--540, 1983.

\bibitem{DBLP:journals/corr/DosovitskiyB15}
Alexey Dosovitskiy and Thomas Brox.
\newblock Inverting convolutional networks with convolutional networks.
\newblock {\em CoRR}, abs/1506.02753, 2015.

\bibitem{Dosovitskiy:2016:GIP:3157096.3157170}
Alexey Dosovitskiy and Thomas Brox.
\newblock Generating images with perceptual similarity metrics based on deep
  networks.
\newblock In {\em Proceedings of the 30th International Conference on Neural
  Information Processing Systems}, NIPS'16, pages 658--666, USA, 2016. Curran
  Associates Inc.

\bibitem{DoubleDIP}
Yossi Gandelsman, Assaf Shocher, and Michal Irani.
\newblock "double-dip": Unsupervised image decomposition via coupled
  deep-image-priors.
\newblock 6 2019.

\bibitem{texturesynth}
Leon Gatys, Alexander~S Ecker, and Matthias Bethge.
\newblock Texture synthesis using convolutional neural networks.
\newblock In C. Cortes, N.~D. Lawrence, D.~D. Lee, M. Sugiyama, and R. Garnett,
  editors, {\em Advances in Neural Information Processing Systems 28}, pages
  262--270. Curran Associates, Inc., 2015.

\bibitem{style_transfer}
Leon~A. Gatys, Alexander~S. Ecker, and Matthias Bethge.
\newblock A neural algorithm of artistic style, 2015.

\bibitem{gan}
Ian Goodfellow, Jean Pouget-Abadie, Mehdi Mirza, Bing Xu, David Warde-Farley,
  Sherjil Ozair, Aaron Courville, and Yoshua Bengio.
\newblock Generative adversarial nets.
\newblock In Z. Ghahramani, M. Welling, C. Cortes, N.~D. Lawrence, and K.~Q.
  Weinberger, editors, {\em Advances in Neural Information Processing Systems
  27}, pages 2672--2680. Curran Associates, Inc., 2014.

\bibitem{resnet}
Kaiming He, Xiangyu Zhang, Shaoqing Ren, and Jian Sun.
\newblock Deep residual learning for image recognition.
\newblock {\em arXiv preprint arXiv:1512.03385}, 2015.

\bibitem{Heusel}
Martin Heusel, Hubert Ramsauer, Thomas Unterthiner, Bernhard Nessler, and Sepp
  Hochreiter.
\newblock Gans trained by a two time-scale update rule converge to a local nash
  equilibrium.
\newblock In {\em Proceedings of the 31st International Conference on Neural
  Information Processing Systems}, NIPS'17, pages 6629--6640, USA, 2017. Curran
  Associates Inc.

\bibitem{noglobal}
Baker Nicholasand~Lu Hongjing, Erlikhman Gennady, and Kellman~Philip J.
\newblock Deep convolutional networks do not classify based on global object
  shape.
\newblock {\em PLOS}, 2018.

\bibitem{huang2017sgan}
Xun Huang, Yixuan Li, Omid Poursaeed, John Hopcroft, and Serge Belongie.
\newblock Stacked generative adversarial networks.
\newblock In {\em CVPR}, 2017.

\bibitem{pix2pix2017}
Phillip Isola, Jun-Yan Zhu, Tinghui Zhou, and Alexei~A Efros.
\newblock Image-to-image translation with conditional adversarial networks.
\newblock {\em The IEEE Conference on Computer Vision and Pattern Recognition
  (CVPR)}, 2017.

\bibitem{gansteerability}
Ali Jahanian, Lucy Chai, and Phillip Isola.
\newblock On the "steerability" of generative adversarial networks.
\newblock {\em arXiv preprint arXiv:1907.07171}, 2019.

\bibitem{Johnson_2016}
Justin Johnson, Alexandre Alahi, and Li Fei-Fei.
\newblock Perceptual losses for real-time style transfer and super-resolution.
\newblock {\em Lecture Notes in Computer Science}, page 694–711, 2016.

\bibitem{adam}
Diederik~P. Kingma and Jimmy Ba.
\newblock Adam: {A} method for stochastic optimization.
\newblock {\em CoRR}, abs/1412.6980, 2014.

\bibitem{alexnet}
Alex Krizhevsky, Ilya Sutskever, and Geoffrey~E. Hinton.
\newblock Imagenet classification with deep convolutional neural networks.
\newblock {\em Communications of the ACM}, 60(6):84--90, jun 2017.

\bibitem{Mahendran_2015}
Aravindh Mahendran and Andrea Vedaldi.
\newblock Understanding deep image representations by inverting them.
\newblock {\em 2015 IEEE Conference on Computer Vision and Pattern Recognition
  (CVPR)}, Jun 2015.

\bibitem{MSGAN}
Qi Mao, Hsin-Ying Lee, Hung-Yu Tseng, Siwei Ma, and Ming-Hsuan Yang.
\newblock Mode seeking generative adversarial networks for diverse image
  synthesis.
\newblock In {\em IEEE Conference on Computer Vision and Pattern Recognition},
  2019.

\bibitem{lsgan}
Xudong Mao, Qing Li, Haoran Xie, Raymond Y.~K. Lau, and Zhen Wang.
\newblock Least squares generative adversarial networks.
\newblock In {\em Computer Vision (ICCV), IEEE International Conference on},
  2017.

\bibitem{sngan}
Takeru Miyato, Toshiki Kataoka, Masanori Koyama, and Yuichi Yoshida.
\newblock Spectral normalization for generative adversarial networks.
\newblock In {\em International Conference on Learning Representations}, 2018.

\bibitem{nguyen2016}
A. {Nguyen}, J. {Yosinski}, and J. {Clune}.
\newblock {Multifaceted Feature Visualization: Uncovering the Different Types
  of Features Learned By Each Neuron in Deep Neural Networks}.
\newblock In {\em Workshop on Visualization for Deep Learning, International
  Conference on Machine Learning (ICML)}, June 2016.

\bibitem{acgan}
Augustus Odena, Christopher Olah, and Jonathon Shlens.
\newblock Conditional image synthesis with auxiliary classifier gans.
\newblock In {\em Proceedings of the 34th International Conference on Machine
  Learning - Volume 70}, ICML'17, pages 2642--2651. JMLR.org, 2017.

\bibitem{dcgan}
Alec Radford, Luke Metz, and Soumith Chintala.
\newblock Unsupervised representation learning with deep convolutional
  generative adversarial networks, 2015.

\bibitem{journals/corr/SimonyanVZ13}
Karen Simonyan, Andrea Vedaldi, and Andrew Zisserman.
\newblock Deep inside convolutional networks: Visualising image classification
  models and saliency maps.
\newblock {\em CoRR}, abs/1312.6034, 2013.

\bibitem{vgg}
Karen Simonyan and Andrew Zisserman.
\newblock Very deep convolutional networks for large-scale image recognition.
\newblock In {\em International Conference on Learning Representations}, 2015.

\bibitem{Teterwak_2019_ICCV}
Piotr Teterwak, Aaron Sarna, Dilip Krishnan, Aaron Maschinot, David Belanger,
  Ce Liu, and William~T. Freeman.
\newblock Boundless: Generative adversarial networks for image extension.
\newblock In {\em The IEEE International Conference on Computer Vision (ICCV)},
  October 2019.

\bibitem{DIP}
Dmitry Ulyanov, Andrea Vedaldi, and Victor Lempitsky.
\newblock Deep image prior.
\newblock {\em arXiv:1711.10925}, 2017.

\bibitem{Upchurch_2017}
Paul Upchurch, Jacob Gardner, Geoff Pleiss, Robert Pless, Noah Snavely, Kavita
  Bala, and Kilian Weinberger.
\newblock Deep feature interpolation for image content changes.
\newblock {\em 2017 IEEE Conference on Computer Vision and Pattern Recognition
  (CVPR)}, Jul 2017.

\bibitem{pix2pixHD}
Ting-Chun Wang, Ming-Yu Liu, Jun-Yan Zhu, Andrew Tao, Jan Kautz, and Bryan
  Catanzaro.
\newblock High-resolution image synthesis and semantic manipulation with
  conditional gans.
\newblock In {\em Proceedings of the IEEE Conference on Computer Vision and
  Pattern Recognition}, 2018.

\bibitem{yang2016highresolution}
Chao Yang, Xin Lu, Zhe Lin, Eli Shechtman, Oliver Wang, and Hao Li.
\newblock High-resolution image inpainting using multi-scale neural patch
  synthesis, 2016.

\bibitem{yosinski-understanding}
Jason Yosinski, Jeff Clune, Anh Nguyen, Thomas Fuchs, and Hod Lipson.
\newblock Understanding neural networks through deep visualization.
\newblock In {\em Deep Learning Workshop, International Conference on Machine
  Learning (ICML)}, 2015.

\bibitem{journals/corr/YosinskiCNFL15}
Jason Yosinski, Jeff Clune, Anh~Mai Nguyen, Thomas~J. Fuchs, and Hod Lipson.
\newblock Understanding neural networks through deep visualization.
\newblock {\em CoRR}, abs/1506.06579, 2015.

\bibitem{SAGAN}
Han Zhang, Ian Goodfellow, Dimitris Metaxas, and Augustus Odena.
\newblock Self-attention generative adversarial networks.
\newblock In Kamalika Chaudhuri and Ruslan Salakhutdinov, editors, {\em
  Proceedings of the 36th International Conference on Machine Learning},
  volume~97 of {\em Proceedings of Machine Learning Research}, pages
  7354--7363, Long Beach, California, USA, 09--15 Jun 2019. PMLR.

\bibitem{zhang2016colorful}
Richard Zhang, Phillip Isola, and Alexei~A Efros.
\newblock Colorful image colorization.
\newblock In {\em ECCV}, 2016.

\bibitem{zhang2018perceptual}
Richard Zhang, Phillip Isola, Alexei~A Efros, Eli Shechtman, and Oliver Wang.
\newblock The unreasonable effectiveness of deep features as a perceptual
  metric.
\newblock In {\em CVPR}, 2018.

\bibitem{places}
Bolei Zhou, Agata Lapedriza, Aditya Khosla, Aude Oliva, and Antonio Torralba.
\newblock Places: A 10 million image database for scene recognition.
\newblock {\em IEEE Transactions on Pattern Analysis and Machine Intelligence},
  2017.

\end{thebibliography}
